\documentclass[letterpaper, 10 pt, journal, twoside]{IEEEtran}

\usepackage{amsmath}
\usepackage[table,xcdraw]{xcolor}
\usepackage{amssymb}
\usepackage{graphicx}
\usepackage[hidelinks]{hyperref}
\usepackage{float} 
\usepackage[ruled,noend]{algorithm2e} 
\newcommand{\mean}[2]{\begin{footnotesize}#1$\pm$#2\end{footnotesize}}
\newcommand{\justmean}[1]{\begin{footnotesize}#1\end{footnotesize}}
\newcommand{\method}[1]{\textit{#1}}

\usepackage{etoolbox}
\makeatletter
\patchcmd{\@algocf@start}
  {-1.5em}
  {0pt}
  {}{}
\makeatother

\begin{document}
%
\title{DALL-E-Bot: Introducing Web-Scale\\Diffusion Models to Robotics}
%
%
%

\author{Ivan Kapelyukh$^{*1,2}$, Vitalis Vosylius$^{*1}$, Edward Johns$^{1}$%
\thanks{Manuscript received: February, 24, 2023; Accepted April 11, 2023.}
\thanks{This paper was recommended for publication by Editor Aleksandra Faust upon evaluation of the Associate Editor and Reviewers' comments.
This work was supported by Dyson Technology Ltd, and the Royal Academy of Engineering under the Research Fellowship Scheme.} 
\thanks{$^*$Ivan Kapelyukh and Vitalis Vosylius are co-first authors.}
\thanks{$^{1}$Ivan Kapelyukh, Vitalis Vosylius and Edward Johns are with the Robot Learning Lab at Imperial College London.
        {\tt\footnotesize {\{ik517,vv19,e.johns\}@imperial.ac.uk}}}%
\thanks{$^{2} $Ivan Kapelyukh is also with the Dyson Robotics Lab at Imperial College London.
        {}}%
\thanks{Digital Object Identifier (DOI): 10.1109/LRA.2023.3272516. © 2023 IEEE.}
}
%
%

\markboth{IEEE Robotics and Automation Letters. Preprint Version. Accepted April, 2023}
{Kapelyukh, Vosylius, Johns: DALL-E-Bot} 

%



\maketitle

\begin{abstract}
We introduce the first work to explore web-scale diffusion models for robotics. DALL-E-Bot enables a robot to rearrange objects in a scene, by first inferring a text description of those objects, then generating an image representing a natural, human-like arrangement of those objects, and finally physically arranging the objects according to that goal image. We show that this is possible zero-shot using DALL-E, without needing any further example arrangements, data collection, or training. DALL-E-Bot is fully autonomous and is not restricted to a pre-defined set of objects or scenes, thanks to DALL-E's web-scale pre-training. Encouraging real-world results, with both human studies and objective metrics, show that integrating web-scale diffusion models into robotics pipelines is a promising direction for scalable, unsupervised robot learning. Videos are available on our webpage at: \textcolor{blue}{\href{https://www.robot-learning.uk/dall-e-bot}{https://www.robot-learning.uk/dall-e-bot}}.
\end{abstract}


%
\IEEEpeerreviewmaketitle

\section{Introduction}
%
%
%
%


\IEEEPARstart{M}{any} everyday tasks, such as setting a dining table, tidying an office, or packing groceries, can be expressed as an object rearrangement problem \cite{rearrangement} in robotics: given a set of objects, determine a goal pose for each and then physically move the objects accordingly. However, calculating these goal poses is a challenging problem, due to the diversity of factors that should be considered. For example, when tidying a room, the created arrangement should be semantically appropriate, aesthetically appealing, physically stable, and convenient for a human to use.

Most prior approaches for predicting goal states (i.e. a goal pose for each object) rely on a training dataset of example arrangements, where objects are placed into desirable poses either manually \cite{jiang-human-context,neatnet,housekeep,kang-simulated-annealing}, or in simulation using a hand-crafted function \cite{structformer,structdiffusion,adaptable-planners}. At test time, the robot can then rearrange a given set of objects into a similar arrangement. This approach can be effective if the training and testing scenes are similar, but it is challenging to scale to unstructured environments such as homes, due to the sheer diversity of objects present, and the combinatorial complexity of acceptable arrangements. Today's methods often still require hundreds or thousands of examples to be provided \cite{housekeep,structformer,structdiffusion,adaptable-planners}.

As an alternative direction to manually collecting datasets of desirable object arrangements, we observe that human preferences for object arrangement are implicit in images of human-arranged scenes, which are abundant at scale on the Web. Extracting arrangement preferences from this web-scale data is therefore an attractive research direction, as this could enable generalisation to a broad set of objects and scenes. Recently, diffusion models such as OpenAI’s DALL-E 2 \cite{dalle2} have been trained on hundreds of millions of image-caption pairs from the Web. These models learn a language-conditioned distribution over natural images, from which new images can be generated given a text prompt. In our work, we show for the first time how these web-scale diffusion models can be used directly in robotics pipelines, without requiring any further training, thus offering an exciting direction towards scalable learning of object rearrangement.

\begin{figure}[t!]
    \centerline{\includegraphics[width=\linewidth]{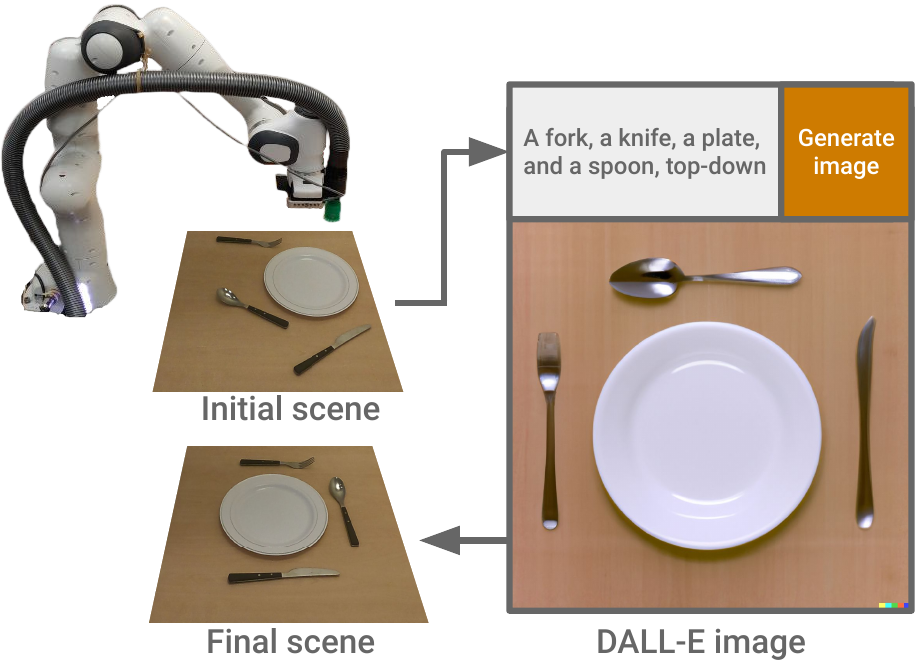}}
    \caption{In DALL-E-Bot, the robot prompts DALL-E with a list of objects it has detected, which then generates an image of a human-like arrangement of those objects. The robot then rearranges the real objects via pick-and-place to match the generated goal image.}
    \label{fig:teaser}
\end{figure}

Our framework, called DALL-E-Bot, uses these pre-trained diffusion models as ``imagination engines'' for robots. As shown in Fig. \ref{fig:teaser}, DALL-E-Bot enables a robot to generate an image of a natural, human-like goal state for a rearrangement task, requiring as input only the image that the robot initially observes of the scene. Consequently, the robot can then determine the goal pose for each object, and execute the rearrangement with pick-and-place actions. The contributions presented in this paper include:

\textbf{(1)} A modular pipeline for performing rearrangement tasks, from visual perception to real-world robot execution, incorporating a web-scale diffusion model. This involves creating an object-centric representation of the scene, without limiting the method to a pre-defined set of classes. We also show empirically that a sample-and-filter strategy when using diffusion models is crucial for performance.
\textbf{(2)} Techniques for crossing the domain gap between real images and diffusion-generated images. Our algorithm combines cross-instance Hungarian matching based on CLIP features, ICP alignment on segmentation masks, and photometric loss on semantic feature maps.
\textbf{(3)} Leveraging the inpainting capability of diffusion models to allow the robot to take into account the poses of objects pre-placed by the human, enabling collaborative human-robot rearrangement.
\textbf{(4)} Experiments using both subjective and objective metrics, including a user study collecting 3000 user ratings, where we evaluate our method on several useful everyday rearrangement tasks.

The DALL-E-Bot framework has several useful properties.
\textbf{(1)~Zero-shot:} it uses only pre-trained models like DALL-E, with no demonstrations or training required. This scalable approach greatly eases the burden on both researchers and users.
\textbf{(2)~Open-set:} it is not restricted to a specific set of objects or scenes, since these diffusion models are pre-trained on web-scale data.
\textbf{(3)~Autonomous:} it does not require any user-provided goal state specification or~supervision.

To the best of our knowledge, this is the first work to investigate web-scale diffusion models for robotics and unlock these advantageous properties.

\section{Related Work}

\subsection{Predicting Goal Arrangements}

We now highlight prior approaches to predicting goal poses for rearrangement tasks. Some methods view the prediction of goal poses as a classification problem, by choosing from a set of discrete options for an object's placement. For house-scale rearrangement, a pre-trained language model can be used to predict goal receptacles such as tables \cite{housekeep}, and out-of-place objects can be detected automatically \cite{tidee}. At a room level, the correct drawer or shelf can be classified \cite{organisational-principles-classification}, taking preferences into account \cite{abdo-shelves}. Lower-level prediction from a dense set of goal poses can be achieved with a graph neural network \cite{efficient-interpretable} or a preference-aware transformer \cite{adaptable-planners}. Our framework generates high-resolution images of where objects should be placed, thus does not require a set of discrete options to be pre-defined, and can predict more precise poses than is possible with language. 

Methods for predicting continuous object poses typically use a dataset of example arrangements. They can learn spatial preferences with a graph VAE \cite{neatnet}. For language-conditioned rearrangement, an autoregressive transformer \cite{structformer} can be used, or a diffusion model over poses can be combined with learned discriminators to avoid collisions \cite{structdiffusion}. For furniture layout generation, there are methods which predict goal states via iterative denoising \cite{legonet}, or avoid collisions during the rearrangement process using gradient fields \cite{targf}. Other rearrangement approaches use full demonstrations \cite{cliport,transporters}, or apply priors such as human pose context \cite{jiang-human-context}. Unlike all of these works, our proposed framework does not require collecting and training on a dataset of rearrangement examples, which often restricts these methods to a specific set of objects and scenes. Instead, we show that exploiting existing web-scale diffusion models enables zero-shot rearrangement.

\subsection{Goal Images for Task Specification}

Many manipulation methods use a goal image to specify the goal state. This includes flow-based rearrangement which generalises to novel objects \cite{ifor}, as well as manipulation policies which are learned from demonstration \cite{lfp} or through reinforcement learning \cite{rl-imagined-goals}. Requiring a provided goal image often places a burden on the user to complete the task themselves in order to show the goal state. Instead, our proposed framework for automatically generating realistic goal images can be used together with all these existing methods to make them truly autonomous by avoiding manual goal specification.

\subsection{Diffusion Models}

Web-scale image diffusion models such as DALL-E are at the heart of our framework. A diffusion model \cite{sohl2015} is trained to remove added noise from a data sample, e.g. an image. By starting from random noise and iteratively applying many learned denoising steps, a new sample can be generated from the learned distribution. In robotics, diffusion models have been trained to learn the distribution over actions for trajectory planning \cite{levine-diffusion} and for visuomotor control \cite{diffusion-policy}. In our work, we show how pre-trained image diffusion models can be used zero-shot. We use DALL-E 2 \cite{dalle2}, but our framework can also be used with other text-to-image models \cite{stable-diffusion}.

\subsection{Image Generation in Robot Manipulation}

In our work, we study how to generate a goal image using a general-purpose web-scale diffusion model (DALL-E). But image generation models have previously been used for robotics in various other ways. For example, learning an image dynamics model can then be used for visual control \cite{deep-visual-foresight,dvd}. And whilst our work is the first to study web-scale diffusion models for robotics, other recent work \cite{cacti} also uses these diffusion models to add distractor objects during training as a form of data augmentation. Subsequent work uses augmentation to generalise pick-and-place policies to novel objects and environments \cite{genaug}, and to automatically select regions for augmentation using text guidance \cite{rosie}. However, these augmentation methods aim to make existing learned controllers more robust and general, whilst our work aims for zero-shot object rearrangement by generating goal images, without requiring prior learned controllers. Recently, text-to-video diffusion models have also been used in robotics by fine-tuning on robot demonstration videos \cite{unipi}.

\section{Method}

\begin{figure*}[h]
    \centerline{\includegraphics[width=1\linewidth]{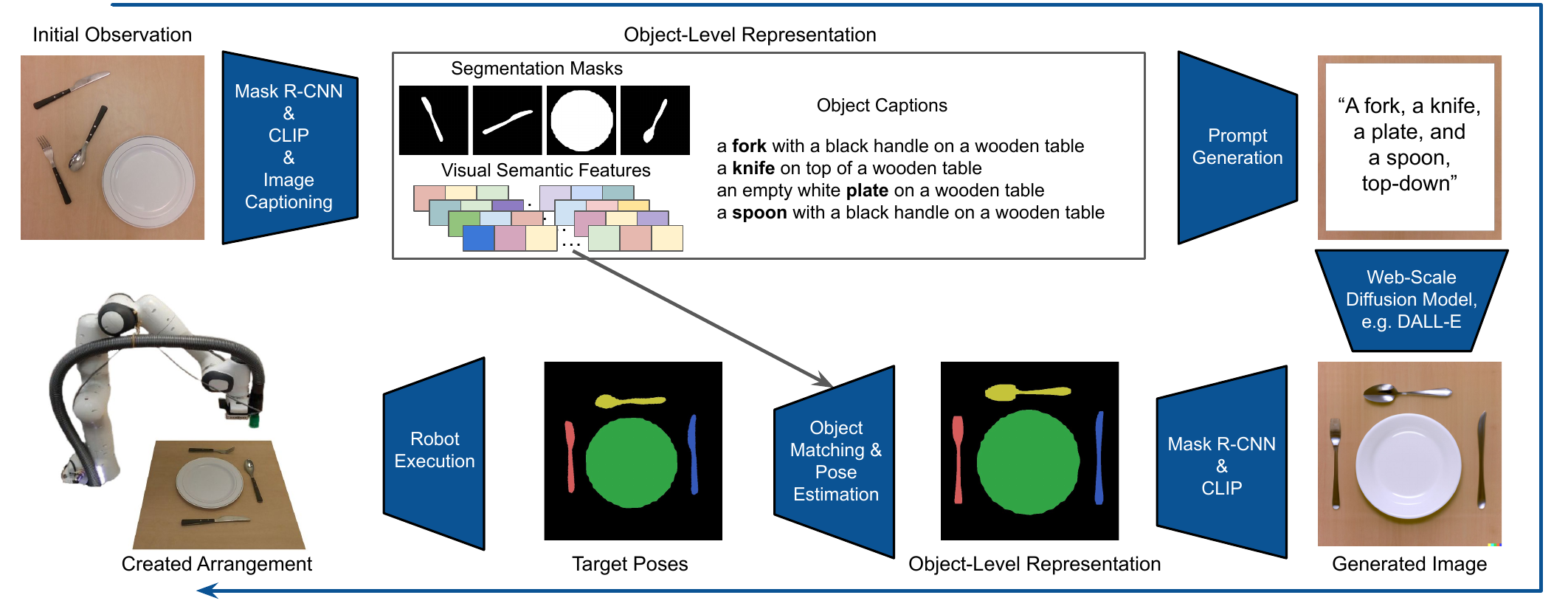}}
    \caption{DALL-E-Bot creates a human-like arrangement of objects in the scene using a modular approach. First, the initial observation image is converted into a per-object description consisting of a segmentation mask, an object caption, and a CLIP visual feature vector. Next, a text prompt is constructed describing the objects in the scene and is passed into DALL-E to create a goal image for the rearrangement task, where the objects are arranged in a human-like way. Then, the objects in the initial and generated images are matched using their CLIP visual features, and their poses are estimated by aligning their segmentation masks. Finally, a robot rearranges the scene based on the estimated poses to create the generated arrangement.
    }
    \label{fig:pipeline}
\end{figure*}

\subsection{Overview}\label{ss:overview}
We address the problem of predicting a goal pose for each object in a scene, such that objects can then be rearranged in a human-like way. We propose to predict goal poses zero-shot from a single RGB image $I_I$ of the initial scene.

To achieve this, we propose a modular pipeline shown in Fig.~\ref{fig:pipeline}.  At the heart of our method is a web-scale image diffusion model DALL-E 2 \cite{dalle2}, which, given a text description $\ell$ of the objects in a scene, can generate a goal image $I_G$, depicting a human-like arrangement of those objects. We can sample many such images for a given text description.
We convert an initial RGB observation into a more relevant object-level representation to individually reason about the objects in the scene. This representation consists of text captions $c_i$ of crops of individual objects (used to construct a text prompt $\ell$) together with their segmentation masks $M_i$, and visual-semantic feature vector $v_i$ acquired using the CLIP model \cite{clip}.
We also convert generated images into object-level representations and select the image that has the same number of objects as the initial scene, and best matches the objects in the initial scene semantically.
Using an Iterative Closest Point (ICP) \cite{icp} algorithm in image space, we then register corresponding segmentation masks to obtain transformations, which are applied to each object to achieve the desired arrangement. Finally, we convert these transformations from image space to Cartesian space using a depth camera observation, and deploy a real Franka Emika Panda robot equipped with a compliant suction gripper to rearrange the scene. Since this method is modular, it will improve as the individual components (e.g. segmentation) improve in the future.

\subsection{Object-Level Representation}\label{ss:object-detection}
 
To reason about the poses of individual objects in the observed scene, we need to convert the initial RGB observation into a more functional, object-level representation. We use the Mask R-CNN model \cite{mask-r-cnn} from the Detectron2 library \cite{detectron2} to detect objects in an image and generate segmentation masks $M_i$. This model was pre-trained on the LVIS dataset \cite{lvis}, which has $1200$ object classes, being more than sufficient for many rearrangement tasks. For each object, Mask R-CNN provides us with a bounding box, a segmentation mask, and a class label. However, we found that whilst the bounding box and segmentation mask predictions are usually high quality and can be used for pose estimation (described in Section~\ref{ss:pose-estimation}), the predicted class labels are often incorrect due to the large number of classes in the training dataset.

As we are using labels of objects in the scene (described in Section~\ref{ss:prompt-generation}) to construct a prompt for an image diffusion model, it is crucial for these labels to be accurate. Therefore, instead of using Mask R-CNN's predicted class labels, we pass RGB crops around each object's bounding box through an OFA image-to-text captioning model \cite{ofa}, to get text descriptions $c_i$ of the objects in the initial scene image. Generally, this approach allows us to more accurately predict object class labels and go beyond the objects in Mask R-CNN's training distribution, and even obtain their visual characteristics such as colour or shape. Finally, we also pass each object crop through a CLIP visual model \cite{clip}, giving each object a 512-dimensional visual-semantic feature vector $v_i$. These features will be used later for matching objects between the initial scene image and the generated image. An alternative approach would be to also infer captions for generated objects and use them for matching, but this would rely on the captioning model's ability to recognise generated objects.

In summary, by the end of this stage, we have converted an RGB observation $I_I$ into an object-level representation $\left(M_i, c_i, v_i\right)$, which represents each object by a segmentation mask, a text caption, and a semantic feature vector.

\subsection{Goal Image Generation}\label{ss:prompt-generation}

We wish to generate images of natural and human-like arrangements, given their text descriptions. To this end, we exploit recent advances in text-to-image generation and web-scale diffusion models, by using the publicly-available DALL-E 2 \cite{dalle2} model from OpenAI. This has been trained on a vast number of image-caption pairs from the Web, and represents the conditional distribution $p_\theta(I_G |\ell, I_M)$. Here, $I_G$ is an image generated by the model, $\ell$ is a text prompt, and $I_M$ is an image mask that can be used to prevent the model from changing the values of certain pixels. Distribution $p_\theta$ includes many images with scenes arranged by humans in a natural and usable way. Therefore, by sampling from this distribution, we can generate images depicting human-like arrangements and create those arrangements in the real world by moving objects to the same poses as in the generated images. The ability to condition this distribution on an image mask $I_M$ lets us handle cases where some objects in the scene should not be moved by the robot.

To generate an image using DALL-E, we must first construct a text prompt $\ell$ describing the scene. To this end, we use object captions from our object-level representation. Although full captions, including visual characteristics, could be used to generate images with objects closely resembling the observed ones, in this work, we only use the nouns describing the object's class and leave including visual characteristics for future work. Thus, we extract the class of each object from the caption of its object crop, i.e. we extract ``apple'' from ``a red apple on a wooden table''. We do this by passing the object captions through the Part-of-Speech tagging model \cite{pos-tagging} from the Flair NLP library \cite{flair}, which tags each word as a noun, a verb, etc. From this list of classes, we construct a prompt that makes minimal assumptions about the scene, to allow DALL-E to arrange it in the most natural way. In this work, our experiments use tabletop scenes, with observations captured by a camera mounted on a robot's wrist pointing downwards towards the table. Therefore, we added a ``top-down'' phrase to the prompt to better align the initial and generated images. As such, an example prompt we use would be ``A fork, a knife, a plate, and a spoon, top-down'' (as in Fig.~\ref{fig:pipeline}).

We use DALL-E's ability to condition distribution $p_\theta$ on image masks in three ways. First, if there are objects in the scene that a robot is not allowed to move, we add their contours to $I_M$. This prevents DALL-E from generating these objects in different poses while still allowing for other objects to be placed on top or in them (e.g. a basket cannot be moved, but other objects can be placed inside it). Second, we add a mask of the tabletop's edges in our scene to $I_M$ to visually ground the generated images. This prevents objects from being placed on the edge of the generated image. Additionally, we found empirically that this makes the generated objects have more appropriate sizes. Third, we subtract segmentation masks of all the movable objects from $I_M$, with enlarged masks to remove their shadows. Removing these shadows from $I_M$ is helpful, as if DALL-E sees shadows of objects in their original poses, it often generates objects in the same poses to fit with those shadows, making it harder to generate novel goal poses.

\begin{figure}[hbt]
    \centerline{\includegraphics[width=1\linewidth]{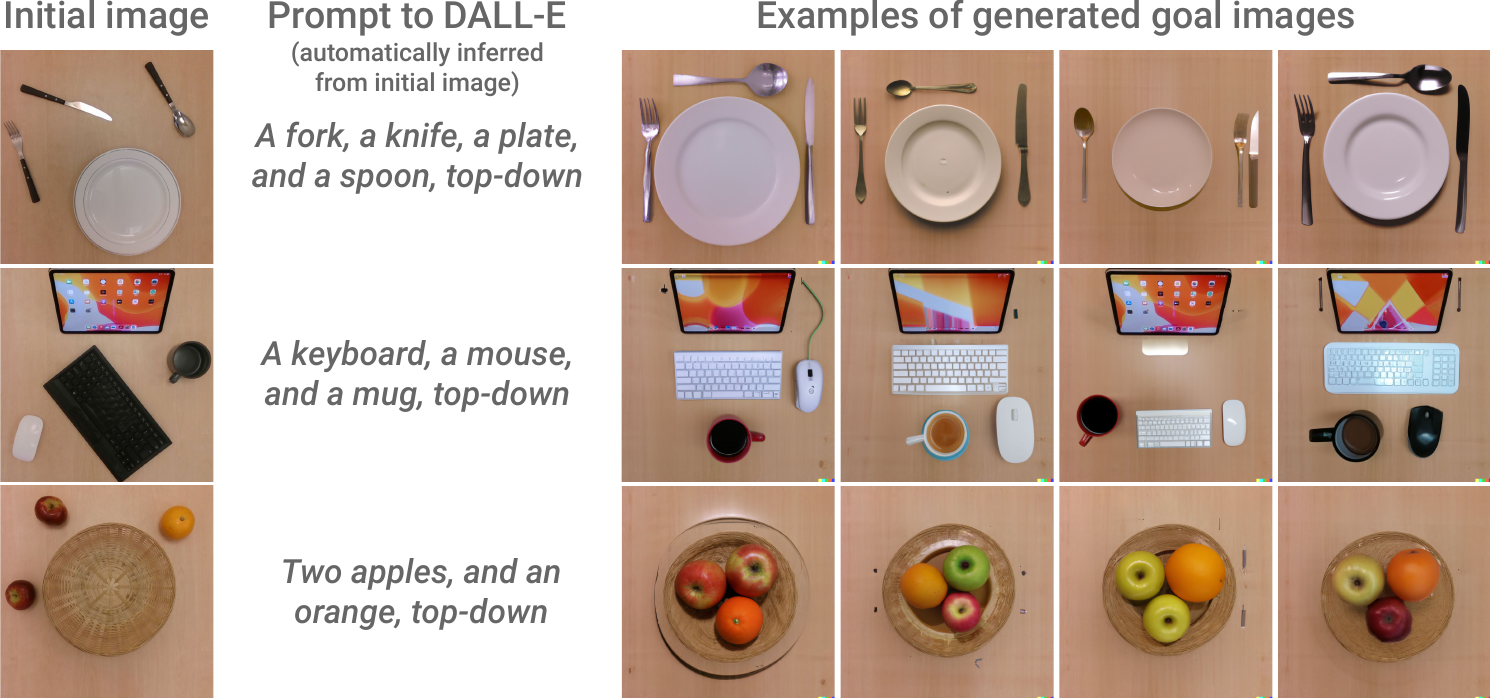}}
    \caption{The robot automatically infers a text prompt $\ell$ from the initial disorganised scene and uses it to generate candidate goal images. DALL-E generates a diverse set of high-quality images depicting human-like arrangements.}
    \label{fig:example-gens}
\end{figure}

Using the prompt $\ell$ and the image mask $I_M$, we sample a batch of images from the conditional distribution $p_\theta(I_G |\ell, I_M)$, representing the text-to-image model. We do so using an automated script and OpenAI's web API. Examples of generated images are shown in Fig. \ref{fig:example-gens}.

\subsection{Image Selection \& Object Matching}\label{ss:object-matching}

In the batch of images generated by DALL-E, not all will be desirable for the rearrangement task; some may have artefacts hindering object detection, others may include extra objects that were not part of the text prompt, etc. Therefore, we need to select the generated image $I_G$ whose objects best match those in the real-world initial image $I_I$.

For each generated image, we obtain segmentation masks and a CLIP semantic feature vector for each object, using the procedure in Section~\ref{ss:object-detection}. Then, we filter out generated images where the number of objects is different to the initial scene, or where movable objects overlap. If there are no generated images which pass these checks, we sample another batch. We then match objects between the generated image and initial image. This is non-trivial since the generated objects are different instances to the real objects, often with a very different appearance. Inspired by \cite{semantic-object-matching}, we compute a similarity score between any two objects (one from $I_I$, and one from $I_G$) using the cosine similarity between their CLIP visual feature vectors. Since greedy matching is not guaranteed to yield optimal results in general, we use the Hungarian Matching algorithm \cite{hungarian-matching} to compute an assignment of each object in the initial image to an object in the generated image, such that the total similarity score is maximised. Then, we select the generated image $I_G$ which has the best overall score with the initial image $I_I$. This image depicts the most similar set of objects to the real scene, and therefore gives the best opportunity for rearranging the real scene.

\subsection{Object Pose Estimation}\label{ss:pose-estimation}

For each object in the initial image, we now know its segmentation mask in the initial image and the corresponding segmentation mask in the generated image. By aligning these masks, we can estimate a transformation from the initial pose (in the initial image) to the goal pose (in the generated image). We rescale each initial segmentation mask, such that the dimensions of its bounding box equal those in the generated image, and then use the Iterative Closest Point (ICP) algorithm \cite{icp} to align the two masks, taking each pixel to be a point. This gives us a 3-DoF $(x, y, \theta)$ transformation $\mathcal{T}$ in pixel space between the initial and goal pose. We run ICP from many random initial poses, due to local optima. For objects with nearly symmetric binary masks such as knives, aligning masks with ICP leads to multiple candidate solutions (for knives, they differ by 180 degrees). To select the correct solution (handle aligned with handle, blade aligned with blade), we pass the generated object image $o_G$ and the transformed real object image $\mathcal{T}\left(o_I \right)$ through a semantic feature map extractor $f_S$ (an ImageNet-trained ResNet \cite{imagenet21k,tresnet}). We select the ICP solution $\mathcal{T}$ which minimises the photometric loss between the semantic feature maps: $\mathcal{L}_S = \left( f_S(o_G) - f_S(\mathcal{T}(o_I)) \right)^2$.

The generated image can depict objects of a different scale than the real objects. Naively moving objects to estimated poses can lead to collisions (if generated objects are smaller) or unnaturally spaced-out arrangements (if generated objects are larger). Therefore, we move objects closer together or further apart based on the mismatch in size, ensuring this does not introduce collisions by moving colliding objects further apart.

Next, we use a depth camera to project the pixel-space poses into 3D space on the tabletop, to obtain a transformation for each object which would move it from the initial real-world pose to the goal real-world pose. Finally, the robot executes these transformations by performing a sequence of pick-and-place operations. We also designed a simple planner which first moves objects that would cause collisions into intermediate poses to the side, before later moving them to their goal poses. More details about the robot execution and hardware used in our experiments can be found in Section~\ref{s:hardware}. Putting these steps together, we summarise our DALL-E-Bot method for autonomous rearrangement in Algorithm~\ref{algo:dallebot}.

\RestyleAlgo{ruled}
\DontPrintSemicolon
\SetInd{0.2em}{0.5em}
\SetKwComment{Comment}{\# }{}
\SetCommentSty{itshape}
\LinesNumbered
\begin{algorithm}
\caption{{\small DALL-E-Bot Autonomous Rearrangement}}\label{algo:dallebot}
Capture image $I_I$ of initial, disorganised scene\;
Get $(M_{Ii}, c_{Ii}, v_{Ii})$ for each obj. $ o_{Ii}$ found in $I_I$\;
\Comment{$M$ from {\small\textsc{MaskRCNN}}, $c$ from {\small\textsc{OFA}}, $v$ from {\small\textsc{CLIP}}}
Construct scene-level text description $\ell$\ from all $c_{Ii}$\;
Sample goal image batch $\{I_G \} \sim p_\theta(I_G |\ell, I_M)$\;
\vspace{0.1cm}
\For{each generated goal image $I_G$}{
Get $(M_{Gj}, c_{Gj}, v_{Gj})$ for each obj. $ o_{Gj}$ found in $I_G$\;
Skip this $I_G$ if fails checks (e.g. wrong obj. count)\;
\Comment{Match objects between $I_I$ and current $I_G$:}
Fill in similarity matrix for each pair $(o_{Ii}, o_{Gj})$: $S_{ij} \gets \cos(v_{Ii}, v_{Gj})$\;
Compute optimal matching using \textsc{{\small{Hungarian}}($S$)}\;
}
Select goal image $I_G$ with max-similarity matches\;
\vspace{0.1cm}
\For{each object $o_{Ii}$ and its match $o_{Gj}$} {
\For{each ICP initialisation}{
Align initial and goal masks: $\mathcal{T}_k \gets$ {\small ICP}($M_{Ii}, M_{Gj}$)\;
\Comment{Get pixelwise loss between semantic feature maps:}
$\mathcal{L}_{Sk} \gets \left( f_S(o_{Gj}) - f_S(\mathcal{T}_k(o_{Ii})) \right)^2$\;
}
Select pose transform for object: $\mathcal{T} \gets \mathrm{argmin}_{\mathcal{T}_{k}}\, L_{Sk}$\;
}
\vspace{0.1cm}
\For{each object $o$ with computed transform $\mathcal{T}$}{
Move other objects away if collision predicted\;
Find grasp and place gripper poses differing by $\mathcal{T}$\;
Execute on robot: \textsc{{\small{PickAndPlace}}}$(o, \mathcal{T})$\;
}
\end{algorithm}

\begin{figure*}[h]
    \centerline{\includegraphics[width=1\linewidth]{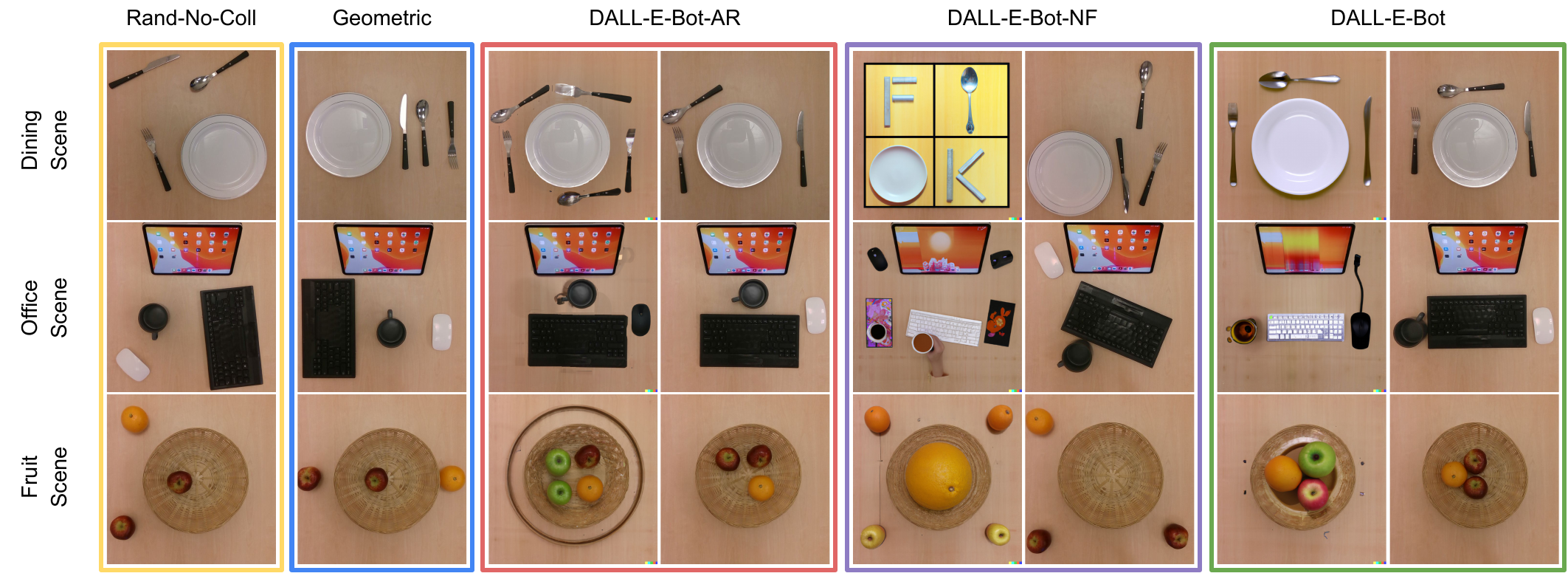}}
    \caption{Examples of scenes rearranged by the robot using different methods. Columns for the methods that use DALL-E include the generated image (left) and the final arrangement (right). For \method{DALL-E-Bot-AR}, images are from the last step.}
    \label{fig:Generated}
\end{figure*}

\section{Experiments}
 
In our experiments, we evaluate the ability of our method to create human-like arrangements using both subjective (Section~\ref{s:qual}) and objective (Section~\ref{s:quant}) metrics.

\subsection{Experiment Setup}\label{s:hardware}

In real-world applications of DALL-E-Bot, users would only see the outcome of the real-world rearrangement, which would include all the errors that might accumulate through the pipeline. To simulate this experience for our evaluation, the predicted arrangements were autonomously created in a 54x54 cm tabletop environment using a 7-DoF Franka Emika robot equipped with a compliant suction gripper to execute a series of top-down pick-and-place operations. For each object, the robot grasps the object in its initial pose and moves it to its target pose, performing a rotation of $\theta$ in between to achieve the target orientation. The robot's motion is calculated using Inverse Kinematics and interpolating Cartesian end-effector poses between a series of waypoints. We use hand-designed grasping primitives to calculate grasping poses for each object, but it is possible to swap in another grasping module such as \cite{graspnet,grasping} into our pipeline if required. The suction gripper is connected to a commercial vacuum device and controlled via an integrated microcontroller. In our experiments, we use a wrist-mounted Intel Realsense D435i RGBD camera and crop images to a 700x700 resolution. After the arrangement is completed, we record the outcome as a top-down RGB image.

\subsection{Zero-Shot Autonomous Rearrangement}\label{s:qual}

First, we explore the following question: \textbf{Can DALL-E-Bot arrange a set of objects in a human-preferred way?} We evaluate on 3 everyday tabletop rearrangement tasks: \textit{dining scene}, \textit{office scene}, and \textit{fruit scene} (Fig.~\ref{fig:Generated}). The dining scene contains four objects: a knife, a fork, a spoon, and a plate. The office scene contains a stationary iPad which the robot is not allowed to move, and three movable objects: a keyboard, a mouse, and a mug. The fruit scene contains a stationary basket, and three movable objects: two apples and an orange.

Since DALL-E-Bot is the first method to predict precise goal poses for rearrangement in a way which is zero-shot (requiring no training arrangements), it cannot be directly compared against methods that require collecting large datasets of example object arrangements, such as~\cite{structformer, structdiffusion}. These existing methods are not designed for the zero-shot setting. Instead, we designed two heuristic-based baselines, which are also zero-shot for a fair comparison. The \method{Rand-No-Coll} baseline places objects randomly in the environment while ensuring they do not overlap. The \method{Geometric} baseline puts all the objects evenly in a straight line such that they are not colliding, and aligns the objects so that they are parallel using their bounding boxes. In addition, we compare our method to two variants. \method{DALL-E-Bot-AR} creates an arrangement in an auto-regressive way, with a sequence of goal images rather than a single image, where each placed object is treated as a stationary object for the next generated image (and thus its contours are added to $I_M$). Here, we do not adjust the poses of the objects based on the size mismatch and do not reject generated images with the wrong number of objects. Finally, \method{DALL-E-Bot-NF} (no filtering) does not filter generated images and always uses the first image. If this image has fewer objects than in the real scene, unmatched objects are placed randomly, whilst avoiding collisions.

Since we aim to create arrangements which are appealing to humans, the most direct evaluation is to ask humans for feedback. This follows related work which also evaluates with human feedback \cite{jiang-human-context,neatnet,housekeep,dalle2,stable-diffusion}, since contrived metrics for arrangement quality may not correlate with what users actually desire. We showed human users images of the final real-world scene created by the robot, and asked them the following question: \textit{``If the robot made this arrangement for you at home, how happy would you be?''}. The user provided a score for each method on a Likert Scale from 1 (very unhappy) to 10 (very happy), while being shown arrangements made by each method side-by-side in a web-based questionnaire. We recruited 40 users representing 18 nationalities, both male and female, with ages ranging from 22 to 71. Each rated the results of 5 methods on 5 random initialisations of 3 scenes, for a total of 3000 ratings. Initialisations were roughly matched for all the methods and all users were shown the same images.

\begin{table}[h]
    \centering
    \begin{tabular}{|l|c|c|c|c|c|}
        \hline
        {\scriptsize Method} & {\scriptsize Dining Scene} & {\scriptsize Office Scene} & {\scriptsize Fruit Scene} & {\scriptsize Mean} \\
        \hline
        {\scriptsize Rand-No-Coll} & \mean{2.03}{1.34} & \mean{3.56}{2.01} & \mean{2.94}{2.01} & \justmean{2.84} \\
        \hline
        {\scriptsize Geometric} & \mean{4.08}{2.27} & \mean{3.36}{2.01} & \mean{3.13}{1.82} & \justmean{3.52} \\
        \hline
        {\scriptsize DALL-E-Bot-NF} & \mean{3.87}{2.78} & \mean{6.54}{2.34} & \mean{7.45}{3.19} & \justmean{5.95} \\
        \hline
        {\scriptsize DALL-E-Bot-AR} & \mean{4.88}{2.61} & \mean{7.37}{2.05} & \mean{9.59}{0.90} & \justmean{7.28} \\
        \hline
        {\scriptsize DALL-E-Bot} & \mean{\textbf{8.01}}{2.03} & \mean{\textbf{7.56}}{2.02} & \mean{\textbf{9.81}}{0.52} & \justmean{\textbf{8.46}} \\
        \hline
    \end{tabular}
    \vspace{0.2cm}
    \caption{User ratings for arrangements by each method. Each cell shows the mean and standard deviation across all users and scene initialisations, with the best in bold.}
    \label{tab:qual}
\end{table}

\begin{figure}[h]
    \centerline{\includegraphics[width=1\linewidth]{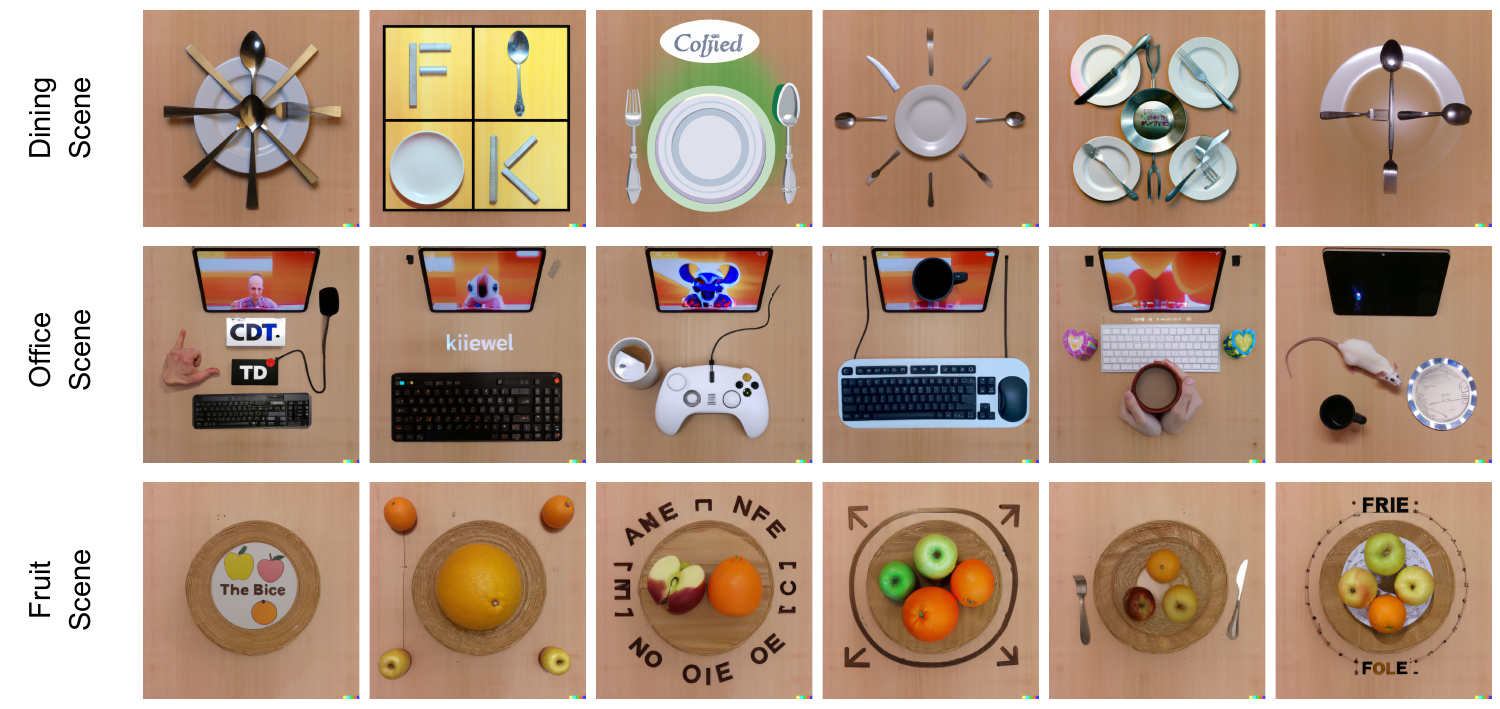}}
    \caption{Diffusion models occasionally produce surreal or unnatural images, such as those shown here. As diffusion models improve, this will happen less frequently. Our sample-and-filter approach removes these and instead selects a suitable goal image, such as those shown in Fig. \ref{fig:example-gens}.}
    \label{fig:no-filter-examples}
\end{figure}

The results of this user study are in Table~\ref{tab:qual}. Example arrangements are shown in Fig.~\ref{fig:Generated}, and videos are available at: \textbf{\textcolor{blue}{\href{https://www.robot-learning.uk/dall-e-bot}{https://www.robot-learning.uk/dall-e-bot}}}. \method{DALL-E-Bot} receives high user scores, showing that it can create satisfactory arrangements zero-shot, without requiring task-specific training. It beats the heuristic baselines, showing that users value semantic correctness for arranging scenes beyond simple geometric alignment, which justifies the use of web-scale learning of these semantic rules. This is especially evident in the dining scene, where DALL-E recognises the semantic structure which can be created from those objects.
The \method{DALL-E-Bot-NF} ablation performs the worst out of the DALL-E-Bot variants on all scenes. This justifies our sample-and-filter approach for using these web-scale models, which ensures that the robot can feasibly create the generated arrangement, rather than naively using the first generated image, which can occasionally be unnatural (see Fig. \ref{fig:no-filter-examples}). The \method{DALL-E-Bot-AR} variant performs well generally but struggles in the dining scene, where the thin cutlery may slip, leading to accumulating error since the method auto-regressively conditions on the objects placed so far. \method{DALL-E-Bot} avoids this issue by jointly predicting all object poses.

\subsection{Placing Missing Objects with Inpainting}\label{s:quant}

In the next experiment, we use objective metrics to answer the following question: \textbf{Can DALL-E-Bot precisely complete a partial arrangement made by a human?} We ask DALL-E-Bot to find a suitable pose for an object that has been masked out from a human-made scene, while the other objects are kept fixed. We study this using the dining scene, because it has the most semantically rigid structure, lending itself well to quantitative, objective evaluation. To create these scenes, we asked ten users (both left and right-handed) the following: ``\textit{Imagine you are sitting down here for dinner. Can you please arrange these objects so that you are happy with the arrangement?}''. As there can be multiple suitable poses for any object, for each of the objects we asked the users to provide any alternative poses that they would also be happy with, while keeping other objects in their original poses. 

Given the image of the arrangement made by a user, we mask out everything except the fixed objects. This means that DALL-E cannot change the pixels belonging to fixed objects. The method must then predict the pose of the missing object. DALL-E-Bot does this by inpainting the missing object somewhere in the image. For a given user, the predicted pose for the missing object is compared against the actual pose in their arrangement. This is done by aligning two segmentation masks of the missing object, one from the actual scene and one at the predicted pose. Since this is for two poses of exactly the same object instance, we find the alignment is highly accurate and can be used to estimate the error between the actual and predicted pose. From this transformation, we take the orientation and distance errors projected into the workspace as our metrics. This is repeated for every object individually as the missing object, and across all the users.

We compare our method to two zero-shot heuristic baselines, \method{Rand-No-Coll} and \method{Geometric}. \method{Rand-No-Coll} places the missing object randomly within the bounds of the image, ensuring it does not collide with the fixed objects. \method{Geometric} first finds a line defined by centroids of segmentation maps of two fixed objects. Then it places the considered object on that line such that it is as close to the fixed objects as possible, does not collide with them, and its orientation is aligned with the orientation of the closest object.

\begin{table}[h]
    \scriptsize
    \centering
\begin{tabular}{|l|c|c|c|c|}
\hline
                    & Fork          & Plate         & Spoon         & Knife        \\ \hline
Method              & cm / deg      & cm / deg      & cm / deg      & cm / deg     \\ \hline
Rand-No-Coll        & 25.85 / 70.32 & 10.78 / - & 27.47 / 42.56 & 23.51 / 99.32 \\ \hline
Geometric & 15.59 / 40.57 & 2.29 / -  & 23.83 / 86.11 & 11.58 / \textbf{1.47} \\ \hline
DALL-E-Bot          & \textbf{4.95} / \textbf{1.26}   & \textbf{1.28} / -  & \textbf{2.13} / \textbf{2.72}   & \textbf{2.1} / 3.27   \\ \hline
\end{tabular}
    \vspace{0.2cm}
    \caption{Position and orientation errors between predicted and user preferred object poses. Each cell shows the median across all users, with the best in bold.}
    \label{tab:quant}
\end{table}

We compare the predicted pose against each of the acceptable poses provided by the user, and report the position and orientation errors from the closest acceptable pose in Table~\ref{tab:quant}. The distribution of acceptable poses is multimodal. Therefore, we present the median error across all users, which is less dominated by outliers than the mean and is a more informative representation of the aggregate performance. DALL-E-Bot outperforms the baselines, and is able to accurately place the missing objects for different users. This implies that it is successfully conditioning on the poses of the other objects in the scene using inpainting, and that the human and robot can create an arrangement collaboratively.


\section{Discussion}

\subsection{Limitations}

\textbf{Top-down pick-and-place}. Our experiments focus on 3-DoF rearrangement, which is sufficient for many everyday tasks. However, future work can extend to 6-DoF object poses with more complex interactions, e.g. to stack shelves. This could draw from recent works on collision-aware manipulation \cite{where-to-start} and learning of skills beyond grasping \cite{johns2021coarse-to-fine}.

\textbf{Overlap between objects}. Currently, our method assumes that movable objects cannot overlap, e.g. the fork cannot go on top of the plate. In future, the robot could plan an order for stacking objects. At the start of the rearrangement, the robot could spread out all the objects on the table to reduce occlusions as it detects all the objects it needs to arrange.

\textbf{Robustness of cross-domain object alignment.} We use ImageNet semantic features, inspired by \cite{fr-slam}, to align real and generated objects. However, the generated objects are sometimes difficult to align, e.g. the generated keyboards lack legible text. As diffusion models improve and with techniques such as \cite{textual-inversion}, this issue will be mitigated.

\subsection{Future Work}

\textbf{Personal preferences}. If objects placed by users are visible in the inpainting mask, DALL-E may implicitly condition images on inferred preferences (e.g. left/right-handedness). Future work could extend to conditioning on preferences inferred in previous scenes arranged by users \cite{neatnet}.

\textbf{Prompt engineering}. Adding terms such as ``neat, precise, ordered, geometric'' for the dining scene improved the apparent neatness of the generated image. As found in other works \cite{lets-think}, there is significant scope to explore this further.

\textbf{Language-conditioned rearrangement}. User instructions can easily be added to the text prompt, e.g. ``plates stacked'' vs ``plates laid out''. Prior work shows that following spatial relations such as ``inside of'' is difficult for some diffusion models \cite{dalle-relations}, but future work could overcome this.

\subsection{Conclusions}

In this paper, we show for the first time that web-scale diffusion models like DALL-E can act as ``imagination engines'' for robots, acting like an aesthetic prior for arranging scenes in a human-like way. This allows for zero-shot, open-set, and autonomous rearrangement, using DALL-E without requiring any further data collection or training. In other words, our system gives web-scale diffusion models an embodiment to actualise the scenes that they imagine. Studies with human users showed that they are happy with the results for everyday rearrangement tasks, and that the inpainting feature of diffusion models is useful for conditioning on pre-placed objects. We believe that this is an exciting direction for the future of robot learning, as diffusion models continue to impress and inspire complementary research communities.

\section*{Acknowledgments}

The authors thank Andrew Davison, Ignacio Alzugaray, Kamil Dreczkowski, Kirill Mazur, Alexander Nielsen, Eric Dexheimer, and Tristan Laidlow for helpful discussions, and Kentaro Wada for developing some of the robot control infrastructure which was used in the experiments.

\ifCLASSOPTIONcaptionsoff
  \newpage
\fi



%
\bibliographystyle{IEEEtran}
\bibliography{IEEEabrv, dallebot}

\end{document}